\pdfoutput=1

\documentclass[11pt]{article}

\usepackage{ACL2023}

\usepackage{times}
\usepackage{latexsym}

\usepackage[T1]{fontenc}

\usepackage[utf8]{inputenc}

\usepackage{microtype}

\usepackage{inconsolata}

\usepackage{natbib}
\usepackage{graphicx}
\usepackage{appendix}
\usepackage{microtype}
\usepackage{bm}
\usepackage{mathtools}
\usepackage{amsfonts,amssymb,amsthm}
\usepackage[ruled,noend]{algorithm2e}
\usepackage{multirow}
\usepackage{enumitem}
\usepackage{soul} 
\usepackage{times}
\usepackage{latexsym}
\usepackage{graphicx}
\usepackage{booktabs}
\usepackage{tcolorbox}
\usepackage{listings}

\usepackage{cleveref}
\crefformat{section}{\S#2#1#3}
\crefformat{subsection}{\S#2#1#3}
\crefformat{subsubsection}{\S#2#1#3}
\crefrangeformat{section}{\S\S#3#1#4 to~#5#2#6}
\crefmultiformat{section}{\S\S#2#1#3}{ and~#2#1#3}{, #2#1#3}{ and~#2#1#3}
\Crefformat{figure}{#2Figure~#1#3}
\Crefmultiformat{figure}{Figures~#2#1#3}{ and~#2#1#3}{, #2#1#3}{ and~#2#1#3}
\Crefformat{table}{#2Table~#1#3}
\Crefmultiformat{table}{Tables.~#2#1#3}{ and~#2#1#3}{, #2#1#3}{ and~#2#1#3}
\Crefformat{appendix}{#2Appx.~\S#1#3}
\crefformat{algorithm}{Alg.~#2#1#3}
\Crefformat{equation}{#2Eq.~#1#3}

\newcommand{\oldmodel}{\textsc{MatCha}}
\newcommand{\model}{\textsc{DePlot}}

\usepackage{color, colortbl}
\usepackage{xcolor}
\definecolor{asparagus}{rgb}{0.53, 0.66, 0.42}
\definecolor{applegreen}{rgb}{0.55, 0.71, 0.0}

\newif\ifcomment\commenttrue
\ifcomment
\newcommand{\pinaforecomment}[3]{\colorbox{#1}{\parbox{.8\linewidth}{#2: #3}}}
\else
\newcommand{\pinaforecomment}[3]{}
\fi

\makeatletter
\def\@fnsymbol#1{\ensuremath{\ifcase#1\or *\or 
   \mathsection\or \mathparagraph\or \|\or **\or \dagger\dagger
   \or \ddagger\ddagger \else\@ctrerr\fi}}
\makeatother

\usepackage{courier}
\usepackage{paralist}

\title{\model: One-shot visual language reasoning \\by plot-to-table translation}

\author{Fangyu Liu$^{\spadesuit \clubsuit}$\thanks{\ \ Work done during Google internship.}\ \ \thanks{\ \ Equal contributions.} \ \ \ Julian Martin Eisenschlos$^{\clubsuit\ast}$ \\ \textbf{Francesco Piccinno$^\clubsuit$ \ \ \ Syrine Krichene$^\clubsuit$ \ \ \ Chenxi Pang$^\clubsuit$ \ \ \ Kenton Lee$^\clubsuit$} \\ \textbf{Mandar Joshi$^\clubsuit$ \ \ \ Wenhu Chen$^\clubsuit$ \ \ \ Nigel Collier$^\spadesuit$ \ \ \ Yasemin Altun$^\clubsuit$} \\
$^\clubsuit$Google DeepMind \ \ \ \ $^\spadesuit$University of Cambridge}

\begin{document}

\maketitle

\begin{abstract}
Visual language such as charts and plots is ubiquitous in the human world. Comprehending plots and charts requires strong reasoning skills. Prior state-of-the-art (SOTA) models require at least tens of thousands of training examples and their reasoning capabilities are still much limited, especially on complex human-written queries. This paper presents the first few(one)-shot solution to visual language reasoning. We decompose the challenge of visual language reasoning into two steps: (1) plot-to-text translation, and (2) reasoning over the translated text. The key in this method is a modality conversion module, named as \model{}, which translates the image of a plot or chart to a linearized table. The output of \model{} can then be directly used to prompt a pretrained large language model (LLM), exploiting the few-shot reasoning capabilities of LLMs. To obtain \model{}, we standardize the plot-to-table task by establishing unified task formats and metrics, and train \model{} end-to-end on this task. \model{} can then be used off-the-shelf together with LLMs in a plug-and-play fashion. Compared with a SOTA model finetuned on thousands of data points, \model{}+LLM with just one-shot prompting achieves a 29.4\% improvement over finetuned SOTA on human-written queries from the task of chart QA.\footnote{Code and models: \href{https://github.com/google-research/google-research/tree/master/deplot}{github.com/google-research/google-research/tree/master/deplot}}\footnote{For questions please contact \texttt{fl399@cam.ac.uk} and \texttt{eisenjulian@google.com}.}
\end{abstract}

\section{Introduction}\label{sec:intro}

Multimodal reasoning on visual language such as plots and charts is an extremely complex task. For downstream tasks such as question answering (QA)
on plots/charts, a model needs to first extract relevant information from the image, organize them in a sensible manner, then perform reasoning over the entries extracted. Previous studies have proposed end-to-end solutions to such methods \citep{lee2022pix2struct,liu2022matcha}. Whilst being an effective solution, end-to-end methods need to be finetuned on large amounts of task data and they still lag behind on queries that require complex reasoning even after finetuning. As an example, the current SOTA model \textsc{MatCha} \citep{liu2022matcha} achieves only 38.2\% accuracy on ChartQA \citep{masry-etal-2022-chartqa} (human written queries). 

In the meantime, large language models (LLMs) such as GPT-3 \citep{brown2020language} and PaLM \citep{chowdhery2022palm} have demonstrated exceptional few-shot reasoning skills, without requiring expensive human annotations. However, it is an open question how multimodal reasoning tasks could benefit from LLMs.
In this work, we propose to decompose the multimodal visual language reasoning problem into: (1) converting the input plot image to a linearized table and (2) passing the linearized table to LLMs for one-shot reasoning. 

\begin{figure*}[!t]
\centering
\includegraphics[width=\linewidth]{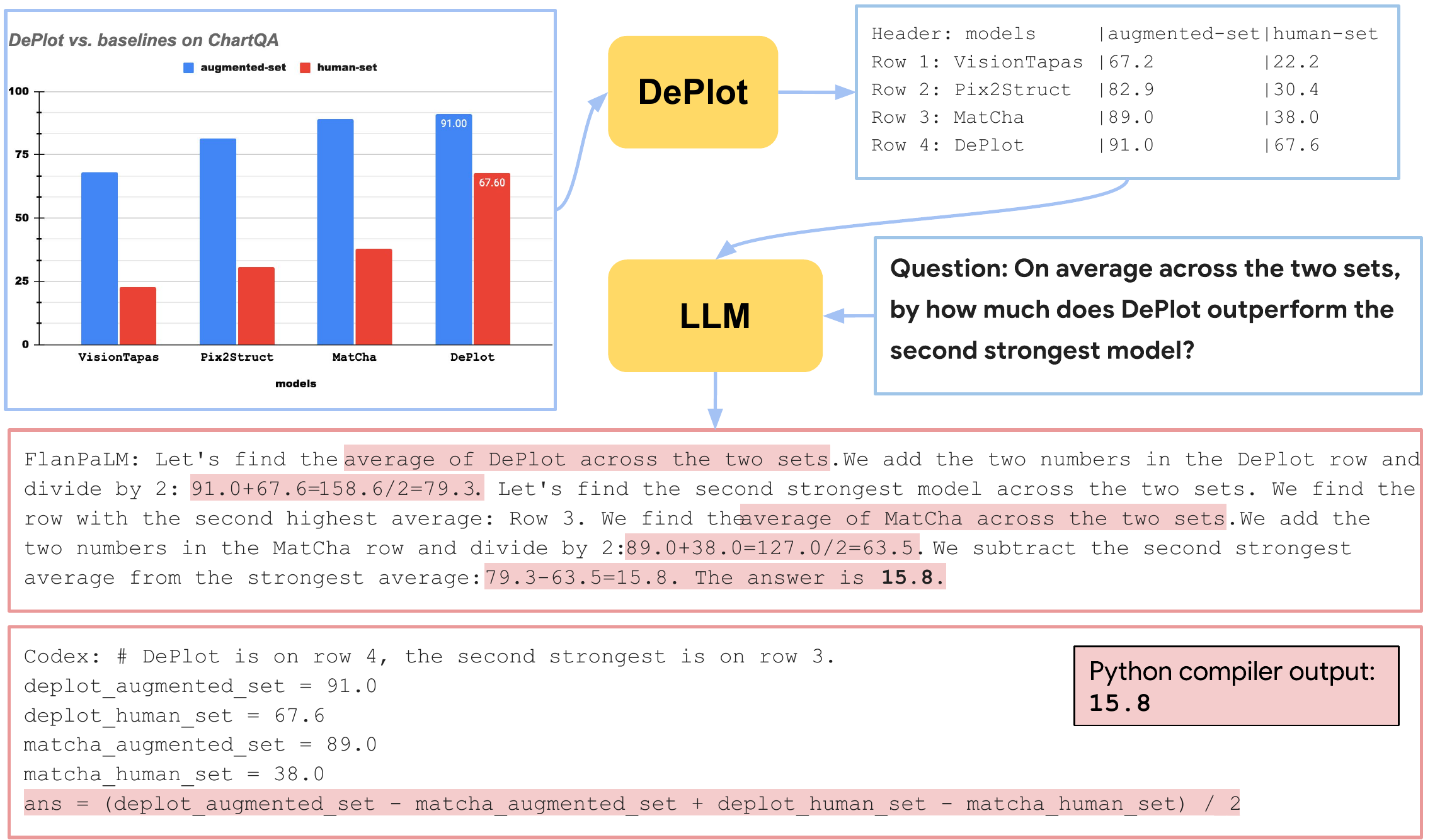}
\caption{An illustration of the \model+LLM method. This is a real example using FlanPaLM~\citep{chung2022scaling} with Chain-of-Thoughts prompting \citep{wei2022cot} and Codex~\cite{chen2021codex} with Program-of-Thoughts prompting~\cite{chen2022program}. The light blue boxes are input (or intermediate forms of the input) to the LLM and the light red box contains the answer generated by the LLMs. Key reasoning steps are highlighted.}
\label{fig:main_fig}
\end{figure*}

The key of the method is 
a modality conversion module called \model{} that maps charts and plots to the underlying data table.
While there has been prior works in chart information extraction, they are usually hybrid systems combining complex hand-designed rules, OCR, keypoint detection, and object segmentation modules \citep{siegel2016figureseer,luo2021chartocr,masry-etal-2022-chartqa}. For different types of charts, distinct approaches have been used \citep{rane2021chartreader,kato2022parsing}.
Besides, there does not exist a unified, consistent, and accurate framework for evaluating chart information extraction -- metrics specific to certain types of charts \citep{siegel2016figureseer} or overly-simplified number matching metrics \citep{luo2021chartocr} have been used.
Our proposed \model{} is an end-to-end image-to-text Transformer model trained with the task of plot-to-table translation. A combination of synthetic and web-crawled charts and plots and their underlying data table are collected as the training corpus. We demonstrate that \model{} significantly outperforms hybrid systems and can uniformly handle all types of charts.
To accurately capture plot-to-table systems' effectiveness (and avoid error propagation to downstream tasks), we propose a novel table matching metric that considers both textual and numeric entries with relative error tolerance, and is invariant to transpositions, row and column permutations.

After accurately translating plot images to texts (as linearized tables), we can pass the output from \model{} in conjunction with a query to LLMs to compute the answer.
We take advantage of novel prompting techniques such as Chain of Thoughts (\emph{CoT})~\cite{wei2022cot}, Self-Consistency (\emph{SC}) ~\cite{wang2022selfconsistency}, and Program of Thoughts (\emph{PoT})~\cite{chen2022program} to elicit more accurate answers. An illustration of the whole process can be seen in 
~\Cref{fig:main_fig}.


To summarize, this work has the following contributions:
(1) We standardize the plot-to-table task and propose a unified and informative metric for table comparison.
(2) We propose a highly-effective  modality conversion model \model{} to translate a multimodal task into a language-only task and then leverage LLMs to solve it with just one shot.
(3) \model+LLM achieves SOTA on ChartQA with just one-shot supervision, outperforming the second best method (which is fully supervised) by 29.4\% on human-written queries.

\section{Background}\label{sec:rw}

\paragraph{Plug-and-play of multimodal pretrained models.} Numerous large pretrained models, either for cross-modal tasks such as CLIP \citep{radford2021learning}, or single-modal tasks, such as GPT-3 and PaLM, have been introduced in the past few years. These pretrained models' strong zero/few-shot inference capabilities have enabled creative solutions to more complex multimodal tasks.
Socratic Models \citep{zeng2022socraticmodels} combine multimodal pretrained models using multimodal prompts for tasks such as multimodal assistive dialogue and robot perception \& planning.
Minds' Eyes \citep{liu2022mind} converts physical reasoning queries into code that could be executed in physical engines.
MAGIC \citep{su2022language} inserts visual control using CLIP in text generation models for unsupervised image captioning.
Similar our work, \citet{yang2022empirical} also translates natural images into texts and leverage GPT-3 for knowledge-based VQA.

However, all above approaches focus on natural images and the tasks of interest usually only require capturing very basic visual information such as types of objects.
Visual language reasoning poses a different set of challenges from natural image reasoning -- it requires, first, accurate and detailed information extraction (IE) from complex visual language data (plots and charts in this work); and secondly very strong numerical reasoning skills to answer queries based on information extracted. While end-to-end fully-supervised models struggle to answer complex human-written queries, \model{} when combined with LLMs can outperform the supervised SOTA by 29.4\%. This is achieved by decomposing the two key challenges in visual language reasoning into leveraging two strong pretrained models that excel at their own respective tasks.

\paragraph{Zero \& few-shot reasoning over tables.} Traditionally, table reasoning tasks are dominated by end-to-end neural models with table-specific architectural designs  \citep{herzig-etal-2020-tapas,yin-etal-2020-tabert,andrejczuk-etal-2022-table}. Recently, there has been a surge in using LLMs to process tables for downstream tasks such as QA. \citet{chen-2023-large} shows that with just one-shot in-context demonstration, GPT-3 could reach near SOTA performance on table QA datasets, on par with end-to-end models trained with at least thousands of training examples. Beyond pure LLM approaches, Binder \citep{cheng2022binding}, Program of Thoughts \citep{chen2022program}, and Program-Aided Language models \citep{gao2022pal} all combine LLMs with compilers/program executors for table reasoning tasks and have achieved SOTA performance. \model{} can be combined with pure LLMs and also any of the aforementioned neural-symbolic methods in a plug-and-play style.

\paragraph{Information extraction from plots and charts.} Prior works on plot/chart IE is usually pipeline-based, combining OCR, object detection/segmentation systems, and hand-crafted rules. Such specialized systems are frequently designed for specific types of graphs, e.g., 
\citet{kato2022parsing} for line graphs, and \citet{rane2021chartreader} for bar plots. ChartBERT \citep{akhtar-etal-2023-reading} adopts an OCR-based method for text extraction from charts and uses two more stages of neural methods for processing the extracted texts. ChartOCR \citep{luo2021chartocr} is a hybrid system that accepts all types of chart inputs and has been adopted by downstream task models for chart QA \citep{masry-etal-2022-chartqa} and summarization \citep{kantharaj-etal-2022-chart}. \model, as an end-to-end neural model, outperforms ChartOCR by very large margins on plot-to-table conversion.

Beyond methodology, the evaluation of plot data extraction tasks has traditionally been ununified. \citet{siegel2016figureseer,luo2021chartocr,kato2022parsing} design different metrics for different types of charts and the metrics can be defined upon coordinates, bounding boxes, or keypoints of the graphs’ objects. However, this measures only the intermediate steps of the data extraction process rather than the quality of data extraction itself. We formulate chart data extraction as a plot-to-table translation task since the ultimate goal of chart IE is obtaining the underlying data table. Besides our work, \citet{masry-etal-2022-chartqa} also considers chart IE as plot-to-table conversion.
However, the metric used in \citet{masry-etal-2022-chartqa} is a number set matching metric, ignoring table structure (i.e., correct organization of the extracted numbers). We propose a better table comparison metric and discuss more in \Cref{sec:chart_to_table_task}.
\section{Standardizing the Plot-to-table Task}\label{sec:chart_to_table_task}

To perform visual language reasoning, we propose to decompose a visual language reasoning task on plots into two steps: (1) converting plots to texts (in the form of linearized tables) using \model{} and (2) inputing the linearized table to LLMs for reasoning.
Accurately performing plot-to-table translation is essential for the downstream visual language reasoning tasks. Plot-to-table is also an important task standalone as it addresses IE from plots/charts, which can benefit applications such as automatic reports and documents digitization.
We will standardize the plot-to-table conversion task in \Cref{sec:chart_to_table_task_def} and propose a new metric for evaluating plot-to-table conversion quality. Then in \Cref{sec:plot_to_text}, we introduce the \model{} model and training procedure for performing plot-to-table conversion.

\subsection{Task Definition}\label{sec:chart_to_table_task_def}

Prior research in table similarity metric is limited. \citet{masry-etal-2022-chartqa} has introduced a metric based on the graph IE metric proposed in \citet{luo2021chartocr}, which we denote \emph{Relative Number Set Similarity} or \texttt{RNSS}. The metric looks only at the unordered set of numeric entries predicted and measures how the predicted set matches the target set of numbers. In the following, we first introduce \texttt{RNSS} more formally and then discuss our rationales of proposing a more well-rounded metric \emph{Relative Mapping Similarity} or \texttt{RMS}.

\paragraph{Relative Number Set Similarity (\texttt{RNSS}).}
Let the model predicted numbers in table be $\mathcal{P} = \{p_i\}_{1 \leq i \leq N}$ and numbers in target tables be $\mathcal{T} = \{t_j\}_{1 \leq j \leq M}$. We compute the pairwise set of relative distances between them:
\begin{equation*}
    \text{D}(p, t) = \min \left(1, \frac{\|p-t\|}{\|t\|}\right).
\end{equation*}
Then the $N \times M$ matrix of distances can be used to find a minimal cost matching between the elements in $\mathcal{P}$ and $\mathcal{T}$, expressed in the form of binary matrix $\mathbf{X}\in \mathbb{R}^{N\times M}$. The final score is computed as
\begin{equation}
    \texttt{RNSS} = 1 - \frac{\sum^N_{i=1}\sum^M_{j=1} \mathbf{X}_{ij}  \text{D}(p_i, t_j)}{\max(N, M)}.
\end{equation}

However, \texttt{RNSS} has several key limitations: it does not distinguish the position of numbers within the table; it completely ignores all non numeric content; it gives credit to very high relative errors; and it does not distinguish precision versus recall losses in table reconstruction.

In contrast, we argue that a metric to measure similarity between tables should satisfy the following desiderata:
\begin{compactenum}
    \item Be invariant to transpositions, as well as permutations of column and rows.
    \item Allow but penalize small errors in numeric or textual values up to a certain threshold.
    \item Clearly reflect losses in precision or recall.
\end{compactenum}

\paragraph{Relative Mapping Similarity (\texttt{RMS}).} In order to address all of these requirements, we propose \texttt{RMS}, which views tables not as sets of numbers but as unordered collection of mappings from row and column headers $(r, c)$ to a single value $v$, which we write $p_i = (p^r_ i, p^c_ i, p^v_ i)$ and $t_j = (t^r_ j, t^c_ j, t^v_ j)$ for each entry in the predicted table $\mathcal{P} = \{p_i\}_{1 \leq i \leq N}$ and the target table $\mathcal{T} =  \{t_j\}_{1 \leq j \leq M}$ respectively.

Following \citet{docvqa}, the distance between textual entries can be measured with \emph{Normalized Levenshtein Distance}, or $\texttt{NL}_\tau$ where the variable $\tau$ is such that values above $\tau$ are set to the maximum of $1$, in order to prevent partial credit for very dissimilar texts. Therefore the distance of two keys $p_i$ and $t_j$ is $\texttt{NL}_\tau\left(p^r || p^c, t^r || t^c\right)$ where $||$ denotes string concatenation. 
The distance between numeric entries is computed using relative distance $\text{D}_\theta (p,t) = \min (1, \|p-t\|/\|t\|)$ and distances above $\theta$ are set to the maximum of $1$.
Combining this two distances we can compute the similarity between two entries in a mapping $\text{D}_{\tau, \theta}(p, t)$ as $
 \left(1-\texttt{NL}_\tau\left(p^r || p^c, t^r || t^c\right)\right) \left(1- \text{D}_\theta\left(p^v, t^v\right)\right)$. When both the keys and values are similar, the similarity $\text{D}_{\tau, \theta}$ is close to $1$ (close to $0$ when dissimilar).

To compute \texttt{RMS}, we first compute the pairwise similarity between keys in $\mathcal{P}$ and $\mathcal{T}$ using the cost function
$1-\texttt{NL}_\tau\left(p^r || p^c, t^r || t^c\right)$. We obtain a similarity matrix with shape $N \times M$ and with the matrix we can identify the minimal cost matching $\mathbf{X} \in \mathbb{R}^{N\times M}$ between the keys (in the form of a binary matrix).  
Then we can compute the precision and recall between two full mappings as the total similarities of the correspondingly matched entries:
\begin{align}
    \texttt{RMS}_{\text{precision}} = 1 - \frac{\sum^N_{i=1}\sum^M_{j=1} \mathbf{X}_{ij}  \text{D}_{\tau, \theta}(p_i, t_j)}{N}, \\
    \texttt{RMS}_{\text{recall}} = 1 - \frac{\sum^N_{i=1}\sum^M_{j=1} \mathbf{X}_{ij}  \text{D}_{\tau, \theta}(p_i, t_j)}{M}.
\end{align}
The $\texttt{RMS}_{\text{F1}}$ score can be computed the harmonic mean of the precision and recall. Because permutations of columns and rows yield the same set of column header, row header, value entries, the resulting metric is invariant to them. In order to allow for table transpositions, we just consider both the table and its transposed version and return the one that corresponds to highest $\texttt{RMS}_{\text{F1}}$ score.

\subsection{Training Plot-to-table Conversion Models}\label{sec:plot_to_text}

Unlike prior works that combine rule-based heuristics, OCR systems, and object/keypoint segmentation/detection systems \citep{siegel2016figureseer,luo2021chartocr,kato2022parsing}, we propose \model{} as an end-to-end solution to plot information extraction. \model{} is conceptually simple yet can robustly work for all types of charts (line, dot, bar, and pie charts) without requiring type-specific engineering and hybrid components. Specifically, we initialize an image-to-text encode-decoder Transformer model with the architecture and weights of the SOTA visual language model \textsc{MatCha} \citep{liu2022matcha}. We continue finetuning the \textsc{MatCha} checkpoint with the task of mapping plots to their underlying data tables. The table is linearized as a textual sequence (markdown format) with \texttt{|} separating cells and \texttt{\textbackslash n} separating rows. \model{} is trained to generate the table from left to right autoregressively.

The training corpus is a set of parallel plot-table pairs collected similar to \citet{liu2022matcha} -- both synthetic data and real world plot-table pairs are combined to form a finetuning corpus.
Specifically, three sources of plot-table pairs are used: (1) synthetic data generated by \citet{liu2022matcha}; (2) synthetic data generated by \citet{methani2020plotqa} (also used in PlotQA dataset); (3) real-world data crawled by \citet{masry-etal-2022-chartqa} (also used in ChartQA). (3) is sourced from four websites, they are \url{statista.com}, \url{pewresearch.com}, \url{ourworldindata.org}, and \url{oecd.org}. The three corpora are mixed with the rate of 1:1:1. The size of each can be seen in \citet{liu2022matcha}.
To avoid data leackage in downstream evaluation, only training set charts from the above datasets are used.
We call our finetuned checkpoint \model.\footnote{Note that the original \textsc{MatCha} model is also pretrained with the task of plot derendering (which includes plot-to-table), however for a different purpose -- i.e., transferring knowledge to downstream finetuning tasks. Our continue finetuning focuses solely on the task of plot-to-table conversion. We also use a much longer sequence length (512 vs. 192) to accommodate long tables.}

\subsection{Human Eval of Plot-to-table Metrics}\label{sec:metric_eval}
To verify that \texttt{RMS} is indeed more sensitive and robust than previously proposed table comparison metric, we conduct human evaluation to compare \texttt{RMS}$_{\text{F1}}$ with the previously used \texttt{RNSS} metric.
Specifically, we sample 50 plot-table pairs where the tables are predictions of the plot-to-table conversion models (to be introduced in more details in \Cref{sec:main_results}).
We score the 50 pairs with \texttt{RNSS} and \texttt{RMS}$_{\text{F1}}$.
Then we collect human judgment of the table prediction quality from 6 human annotators on the 50 examples.\footnote{The 6 annotators are all experienced NLP researchers in information extraction with at least a Master's degree.}
For each instance, the human annotators are given a plot, the model's predicted table, and three questions regarding different aspects of the quality of the predicted table.
The three questions are (1) ``Does the model overgenerate columns/rows or some rows/columns are missing?'', (2) ``Are the x, y label/index names, and title correct?'', and (3) ``Are numbers close to the true values and associated with the correct column, row labels/indexes?''. Annotators should rate the table from 1--5 (the higher the better). We attach the full annotation form in \Cref{sec:human_eval_form}.
The final human score for a plot-table pair is the average of the scores across the three questions across all human annotators.
We compute the Pearson's $r$ and Spearman's $\rho$ correlations between metric scores and human scores.
As shown in \Cref{tab:metric_eval}, under both correlation metrics, we can observe a great improvement of \texttt{RMS}$_{\text{F1}}$ over the baseline \texttt{RNSS}, suggesting that \texttt{RMS}$_{\text{F1}}$ is a much more sensitive and informative metric for evaluating the model generated tables.

\begin{table}[t!]
    \centering
  \scalebox{0.88}{
    \begin{tabular}{lcc}
    \toprule
 Metric & \texttt{RNSS} & \texttt{RMS}$_{\text{F1}}$  \\
    \midrule
Pearson's $r$ & 0.46 & \textbf{0.87} \\
Spearman's $\rho$ & 0.84 & \textbf{0.96} \\

\bottomrule
\end{tabular}
}
    \caption{Correlations between human judgments and metric scores (both \texttt{RNSS} and \texttt{RMS}$_{\text{F1}}$).}
    \label{tab:metric_eval}
\end{table}

\section{Prompting LLMs for Reasoning}\label{sec:prompting_llm}

With \model{} introduced in \Cref{sec:chart_to_table_task}, we can convert a given chart/plot into its textual form (as a linearized table). We can then construct textual prompts by concatenating the linearized tables and the questions for QA tasks.
We follow the typical in-context learning paradigm to prepend a one-shot example before the current prompt.

The full prompts use either Chain-of-Thoughts (CoT) \citep{wei2022cot} or Program-of-Thoughts (PoT) \citep{chen2022program} and can be seen in \Cref{sec:cot_prompt}. They are slightly modified versions of the ones used by ~\citet{chen-2023-large} and \citet{chen2022program} for evaluating reasoning on tabular data. Besides CoT prompting, we also explore combining \model{}+LLM with self-consistency (SC) \citep{wang2022selfconsistency}, which samples a diverse set of reasoning paths and choose the majority-voted answer instead of relying on one greedily-decoded answer as in CoT. In order to simplify performing arithmetic on large numbers, we also tested prompting the models to generate python code that can be passed through an interpreter. In order to do that, we adapt the paradigm from ~\citet{chen2022program,gao2022pal} to the context of tables.
Future work could alternatively take advantage of finetuned tabular QA models such as ~\citet{herzig-etal-2020-tapas} or use LLMs that generate SQL programs~\cite{cheng2022binding} and might require multiple LLM iterative invocations to perform different atomic operations.
\section{Experiment}\label{sec:exp}

We introduce the experimental setup in \Cref{sec:setup} and then the results in \Cref{sec:main_results} including both plot-to-table translation and downstream QA tasks.

\subsection{Experimental Setup}\label{sec:setup}

\paragraph{Training and inference.} \model{} is trained for 10k steps with a maximum sequence length of 512. The other hyperparameters are identical to \textsc{MatCha} pretraining as introduced in \citet{liu2022matcha}. In \model{} inference we set temperature to $0$ (so the output is deterministic). For LLM prompting, in all cases we use temperature of $0.4$.

\paragraph{Datasets and metrics.} We evaluate on two chart/plot question answering benchmarks ChartQA \citep{masry-etal-2022-chartqa} and PlotQA \citep{methani2020plotqa}. ChartQA contains two sets: augmented (aug.) and human where the augmented set is synthetically generated and the human set is human written. Human written queries usually are more diverse and complex, requiring more reasoning while synthetic questions are usually highly templatic. PlotQA is purely synthetic. It contains v1 \& v2 sets where v1 is mostly extractive questions and v2 focuses more on numerical reasoning.  
Both \texttt{RNSS} and \texttt{RMS}$_{\text{F1}}$ are used for evaluating plot-to-table translation (though we have argued that \texttt{RMS}$_{\text{F1}}$ is the more informative metric). Following \citet{masry-etal-2022-chartqa,methani2020plotqa}, exact match accuracy with 5\% tolerance on numerical error is used to report all QA numbers.
        
We list data statistics of plot-to-table training in \Cref{tab:deplot_stats}.  Note that the plot-table pairs are only from ChartQA and PlotQA training sets (not their validation/test sets). The statistics of PlotQA and ChartQA test data are listed in \Cref{tab:stats}. Note that we are also using plot-table pairs from the PlotQA test set for evaluating the plot-to-table task (plot-table pairs from v1 and v2 are identical).

\begin{table}[ht]
    \centering
  \scalebox{0.9}{
    \begin{tabular}{llcc}
    \toprule
 Component & Rate & Size  \\
    \midrule

synthetic (by us) & 33.3\% & 270K \\
ChartQA &  33.3\% & 22K \\
PlotQA & 33.3\% & 224K \\

\bottomrule
\end{tabular}
}
    \caption{Data statistics for the training data of plot-to-table task.}
    \label{tab:deplot_stats}
\end{table}

\paragraph{Hardware.} We train and evaluate our models using 64 GCP-TPUv3. The training of \model{} can be completed in roughly 5 hours.

\paragraph{Parameters.} \model{} has 282M parameters. FlanPaLM has 540B parameters. Codex and GPT3 have roughly 175B parameters.

\begin{table}[t!]
    \centering
  \scalebox{0.9}{
    \begin{tabular}{lcc}
    \toprule
 Dataset & \# Tables & \# QA Pairs  \\
    \midrule
ChartQA (Human) & 625 & 1,250 \\
ChartQA (Machine) & 987 & 1,250 \\
PlotQA (v1) & 33K & 1.2M \\
PlotQA (v2) & 33K & 4.3M \\
\bottomrule
\end{tabular}
}
    \caption{Dataset statistics of the test sets for ChartQA and PlotQA.}
    \label{tab:stats}
\end{table}

\subsection{Main Results}\label{sec:main_results}

\paragraph{Plot-to-table translation.} We evaluate plot-to-table conversion against an OCR and keypoint detection based system proposed by \citet{luo2021chartocr} called ChartOCR. This system also relies on multiple hand-crafted rules that depend on the type of chart. We also compare against two PaLI models~\cite{chen2022pali} (of different input resolutions) finetuned with the same plot-to-table corpus as \model. Finally, we compare with the \textsc{MatCha} base model off-the-shelf. The results are shown in \Cref{tab:plot_to_table}.

\begin{table}[ht]
    \centering
  \scalebox{0.88}{
    \begin{tabular}{lcc}
        \toprule
        Model$\downarrow$, Metric$\rightarrow$ & \texttt{RNSS} & \texttt{RMS}$_{\text{F1}}$ \\
        \midrule
        ChartOCR \citep{luo2021chartocr} & 81.0 & 60.1 \\
        \midrule
        PaLI-17B (res. 224) + plot-to-table & 77.2 & 24.8 \\
        PaLI-17B (res. 588) + plot-to-table  & 90.5 & 74.9 \\
        \textsc{MatCha} \citep{liu2022matcha} & 95.4 & 92.3 \\
        \midrule
        \model{} & \textbf{97.1} & \textbf{94.2} \\
        \bottomrule
    \end{tabular}
    }
    \caption{Benchmarking plot-to-table conversion accuracy on the PlotQA dataset (all individual plots in PlotQA test sets). Both a pipeline-bsed based method (ChartOCR) and end-to-end methods (PaLI-17B and \textsc{MatCha}) are used as baselines. \texttt{RMS}$_{\text{F1}}$ can capture the shortcomings of baselines such as ChartOCR with much greater sensitivity.}
    \label{tab:plot_to_table}
\end{table}

On both metrics, \model{} outperforms the baseline ChartOCR by very significant margins. The gap is especially large on \texttt{RMS}$_{\text{F1}}$ since ChartOCR might suffice to extract numbers from the plot but can struggle to organize the extracted numbers into a structured table with the correct row and column labels. When compared against PaLI and \textsc{MatCha}, \model{} is also better, suggesting that a visual-language-specific architecture/initialization and task-specific finetuning can both boost plot-to-table accuracy. It is also worth noting that PaLI-17B (res. 588) performs much better than the 224-resolution variant, indicating that high input resolution is a key ingredient for chart information extraction.

\begin{table*}[!ht]
    \centering
  \scalebox{0.88}{
    \begin{tabular}{llcccccccccccc}
    \toprule
& & \multicolumn{3}{c}{ChartQA} & & \multicolumn{3}{c}{PlotQA} \\ 
     \cline{3-5}      \cline{7-9}   
 & Model$\downarrow$ &  aug. \colorbox{blue!30}{$\ $} & human \colorbox{red!30}{$\ $} & avg. & & v1 \colorbox{blue!30}{$\ $} & v2 \colorbox{blue!30}{$\ $} & avg.\\  
\midrule
& human ceiling & - & - & - & & - & - & 80.5 \\
\midrule
  \multirow{8}{*}{\rotatebox[origin=c]{90}{Fully-supervised}} & CRCT &  - & - & - & & 76.9 & 34.4 & 55.7 \\
& VL-T5-OCR  & - & - & 41.6 & & 75.9 & 56.0 & 66.0 \\
& T5-OCR & - & - & 41.0 & & 72.6 & 56.2 & 64.4 \\
& VisionTapas-OCR & - & - & 45.5 & & 65.3 & 42.5 & 53.9 \\
& PaLI-17B (res. 224) & 6.2  & 12.6 & 9.4 & & 56.9 & 13.1 & 35.0 \\ 
& PaLI-17B (res. 588) & 64.9 & 30.4 & 47.6 & & 64.5 & 15.2 & 39.8\\ 
& Pix2Struct & 81.6 & 30.5 & 56.0 & & 73.2 & 71.9 & 72.5\\ 
& \oldmodel{}  & 90.2 & 38.2 & 64.2 & & \textbf{92.3} & \textbf{90.7} & \textbf{91.5}\\ 
\midrule 
 \multirow{4}{*}{\rotatebox[origin=c]{90}{One-shot (OURS)}}
 & \model{}+GPT3 CoT  & 37.3 & 36.5 & 36.9 &    & 31.9 & 51.3 & 41.6 \\ 
& \model{}+GPT3 SC  & 42.6 & 41.9 & 42.3 &    & 35.0 & 51.6 & 43.3 \\ 
& \model{}+FlanPaLM CoT & 76.7 & 57.8 & 67.3 &    & 51.3 & 44.9 & 48.1 \\ 
& \model{}+FlanPaLM SC & 78.8 & 62.2 & 70.5 &   & 57.8 & 50.1 & 53.9 \\ 
& \model{}+Codex PoT SC & \textbf{91.8} & 61.6 & 76.7 &   & 58.8 & 69.8 & 64.3 \\
& \model{}+FlanPaLM+Codex PoT SC & 91.0 & \textbf{67.6} & \textbf{79.3} &   & 62.2 & 71.0 & 66.6 \\
\bottomrule 
\end{tabular}
}


\caption{Main experimental results on two plot/chart QA benchmarks ChartQA \& PlotQA. ``\colorbox{red!30}{$\ $}'' denotes human-written queries while ``\colorbox{blue!30}{$\ $}'' denotes synthetic queries.
Detailed introduction of the baselines can be found in \Cref{sec:baselines}. The last six rows show \model+LLM results -- the only one-shot setup. CoT denotes chain-of-thought prompting, SC denotes self-consistency, PoT denotes program-of-thought prompting. The best results are achieved for ChartQA by majority voting across joint 10 CoT and 10 PoT predictions.}
\label{tab:main}
\end{table*}

\paragraph{Downstream tasks.} We list the main results on ChartQA \citep{masry-etal-2022-chartqa} and PlotQA \citep{methani2020plotqa}
in \Cref{tab:main}.
We evaluate different \model+LLM setups. We evaluate chain-of-thoughts (CoT)~\citep{wei2022cot} prompts for GPT-3~\cite{brown2020language} (\texttt{text-davinci-003}) and FlanPaLM~\cite{chung2022scaling} (540B). In addition, we use self-consistency (SC)~\cite{wang2022selfconsistency} across 10 predictions. Finally, we use program-of-thoughts (PoT)~\cite{chen2022program} to prompt Codex~\cite{chen2021codex} (\texttt{code-davinci-002}) to generate python snippets that can be subsequently executed by an interpreter to extract an answer.\footnote{We also evaluated PaLM and FlanPaLM for code generation but found Codex to be more likely to write correct code instead of do the reasoning in comment blocks.} Since some reasoning operations are better done by plain language (like computing an argmax) and some by code snippets (like floating point arithmetic), we find optimal results by doing self-consistency across both CoT and PoT predictions.

\model+LLM performs especially strong on the ChartQA human set (denoted with ``\colorbox{red!30}{$\ $}'') which contains complex human written queries. Compared with prior SOTA \textsc{MatCha}, \model+LLM when combined with FlanPaLM and Codex and Self-Consistency (SC) achieves an improvement of 29.4\% (38.2\%$\rightarrow$67.6\%). This is also the best setup for PlotQA. On the heavily synthetic queries from PlotQA v1 and v2 (denoted with ``\colorbox{blue!30}{$\ $}''), \model+LLM models underperform the end-to-end SOTA \textsc{MatCha}.

In summary, \model+LLM significantly outperforms finetuned SOTA on human-written chart QA queries and overall underperforms finetuned SOTA on synthetic QA queries.
We believe it is especially important to achieve good performance on the human set as it is much more diverse and reflects the real-world challenges. The results suggest \model+LLM's strong capability in solving novel human queries unseen in demonstration.
It is also worth emphasizing again that \model+LLM requires much less supervision than the finetuned SOTA methods (one shot vs. tens of thousands of training examples). We will discuss why \model+LLM underperforms on PlotQA in error analysis (\Cref{sec:case_study}). 

Besides one-shot learning, we also experimented with zero- and few-shot inference. We found the models generally fail without demonstration and few-shot has similar performance as one-shot. After the submission of this paper, we experimented with RLHF-ed LLMs such as ChatGPT\footnote{\href{https://openai.com/blog/chatgpt}{openai.com/blog/chatgpt}}, GPT-4 \citep{OpenAI2023GPT4TR}, and Bard\footnote{\href{https://bard.google.com/}{bard.google.com}}, finding that such aligned conversational models are capable of processing the \model-generated tables in a zero-shot manner. This can potentially further improve \model+LLM's performance on academic benchmarks by large margins.


\section{Analyses and Discussions}\label{sec:analyses}

In this section, we first conduct case studies and error analysis in \Cref{sec:case_study} to better understand \model{}' wins and losses against end-to-end methods. Then in~\Cref{sec:ood} we study the performance of \model{} when exposed to out-of-distribution web charts and plots.

\subsection{Case Study and Error Analysis}\label{sec:case_study}

To more concretely demonstrate the strengths and weaknesses of \model+LLM, we present two case studies for the downstream task ChartQA. We compare \model+FlanPaLM using either PoT or CoT.

First, in \Cref{tab:case_study_m1} we show an example demonstrating the benefit of using LLM and prompting techniques for stronger numerical reasoning. While the finetuned SOTA \textsc{MatCha} wrongly predicts the answer, \model+FlanPaLM (using either CoT or PoT) produces accurately the answer.

\begin{table}[ht!]
  \centering
  \small
  \begin{tabular}{c}
    \begin{minipage}{.46\textwidth}
      \frame{\includegraphics[width=\linewidth, height=56mm]{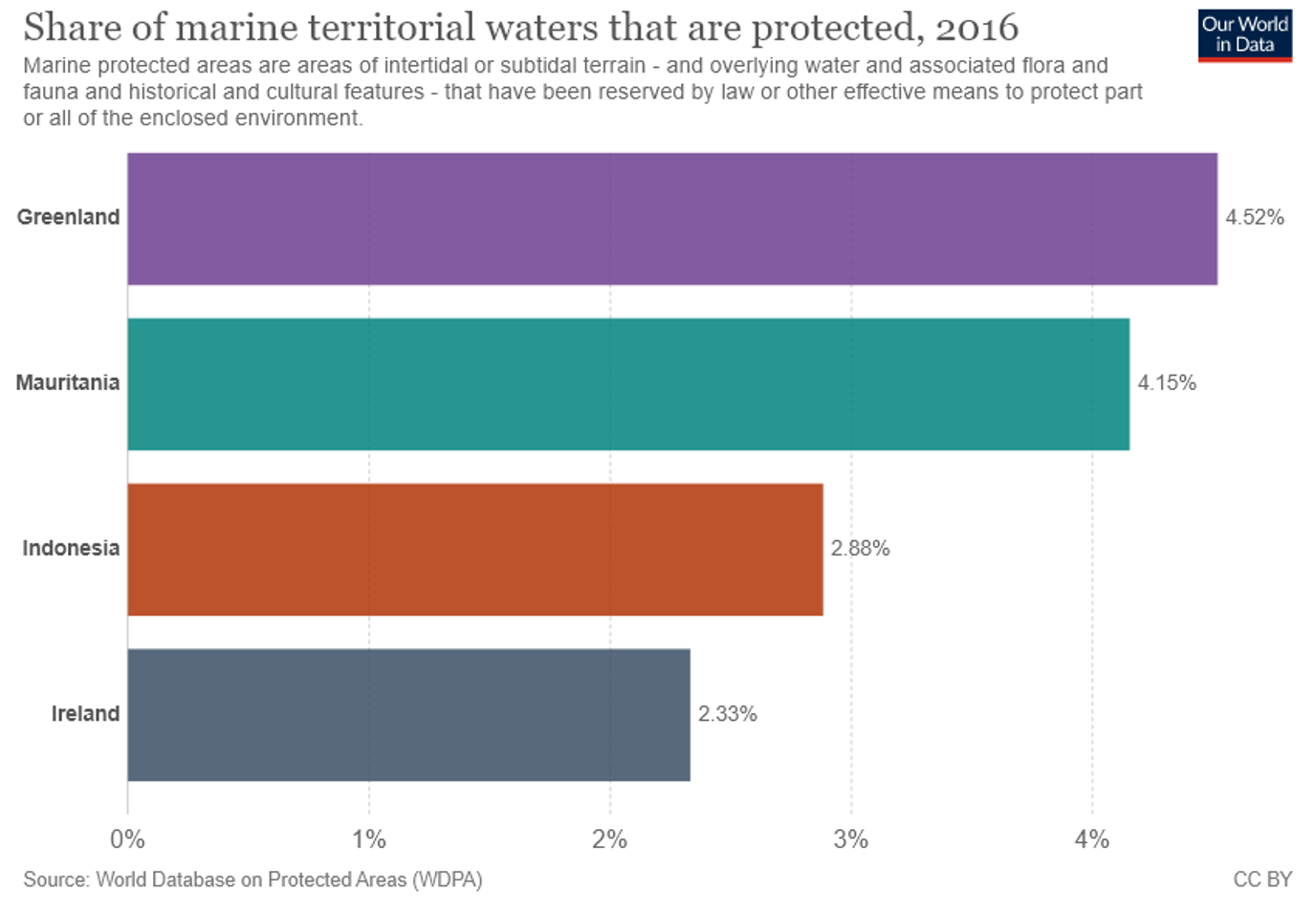}}
    \end{minipage} 
    \\ \\
    \begin{minipage}[t]{.46\textwidth}
    \textit{\textit{Question:} What is the difference between the sum shares of Indonesia and Ireland and the share of Mauritania?} \\ \\
    \textsc{MatCha}: \textcolor{red}{\textbf{0.07}} \\ \\
    \model+FlanPaLM CoT: Let's find the sum of shares of Indonesia and Ireland: 2.88+2.33=5.21. Let's subtract the share of Mauritania: 5.21-4.15=1.06. The answer is  \textcolor[rgb]{0.4,0.8,0}{\textbf{1.06}}.\\ \\
    \model+Codex PoT:\begin{lstlisting}[language=Python]
indonesia = 2.88
ireland = 2.33
mauritania = 4.15
ans=(indonesia+ireland)-mauritania
    \end{lstlisting}
Compiler output: \textcolor[rgb]{0.4,0.8,0}{\textbf{1.06}}
    \end{minipage}
  \end{tabular}
  \caption{\model{}+LLM shows much stronger numerical reasoning skills than the finetuned \textsc{MatCha}.}\label{tab:case_study_m1}
\end{table}

Second, we show an example where the \model+LLM framework fails in \Cref{tab:error}. The LLMs are unable to accurately identify the ``highest value of the gray bar'' since they do not have information about the color of bars. In \Cref{tab:error}, though \model+FlanPaLM correctly predicted ``Yes'', it is correct for the wrong reason -- FlanPaLM randomly chose the highest value in light blue bars which also happens to be smaller than the average of ``identity theft''. This is a typical failure mode where the query refers to a visual attribute but such attribute is lost in plot-to-table translation. In future work, we plan to develop a table encoding scheme that also considers visual attributes to avoid such errors.

\begin{table}[ht!]
  \centering
  \small
  \begin{tabular}{c}
    \begin{minipage}{.46\textwidth}
      \frame{\includegraphics[width=\linewidth, height=76mm]{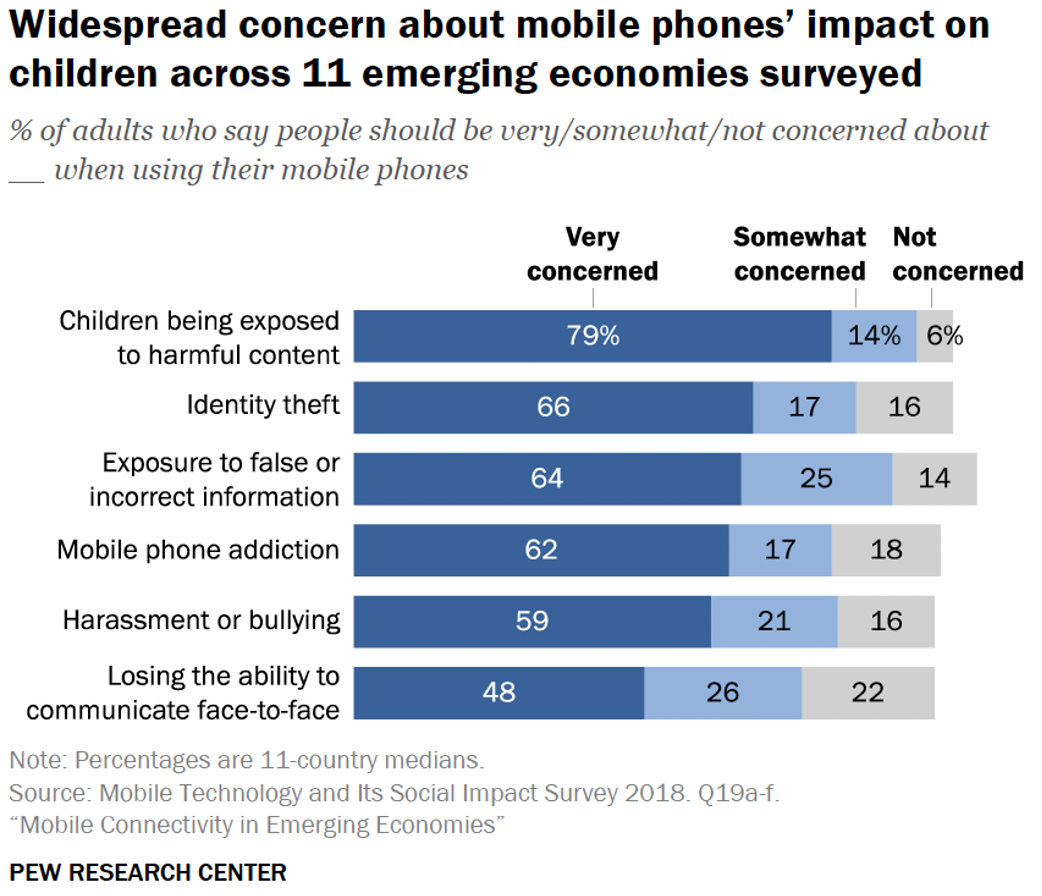}}
    \end{minipage} 
    \\ \\
    \begin{minipage}[t]{.46\textwidth}
    \textit{\textit{Question:} Is the average of all the bars in ``identity theft'' greater than the highest value of the gray bar?} \\ \\
  \textsc{MatCha}: \textcolor{red}{\textbf{No}} \\ \\
  \model+FlanPaLM CoT: Let's find the average of all the bars in "Identity theft". The sum of all the bars is 66+17+16 = 99. The average is 99/3 = 33. The highest value of the gray bar is 26. 33 is greater than 26. The answer is \textcolor[rgb]{0.4,0.8,0}{\textbf{Yes}}. \\ \\
\model+Codex PoT:
\begin{lstlisting}[language=Python]
#Identity theft corresponds to row 5
#Numbers on row 5 are [66, 17, 16]
#Highest value of the gray bar is 79
ans = 66 > 79
\end{lstlisting}
Compiler output: \textcolor{red}{\textbf{No}} 
\end{minipage}
  \end{tabular}
  \caption{\model+LLM struggles with queries related to visual attributes such as color (which is lost in plot-to-table translation).}\label{tab:error}
\end{table}



While \model+LLM has surpassed finetuned SOTA on ChartQA, we notice that the picture on PlotQA is different -- \model{} underperforms finetuned SOTA \textsc{MatCha} by a large margin (66.6\% vs. 91.5\%). Through error analysis, we observe that there are two major reasons. First, synthetic queries are highly templatic and covers only restricted types of questions. Models finetuned with thousands of examples can learn to solve such templatic questions, even better than humans do (human ceiling on PlotQA is just 80.5\% compared with \textsc{MatCha} performance of 91.5\%). However, \model+LLMs only learn from one-shot in-context example and thus cannot exploit such bias encoded in the training set. 
The second reason is the loss of information in plot-to-table translation. Synthetic queries are usually highly extractive and include questions asking visual attributes such as color, shape, or direction of objects in a plot. When plots are converted to tables, such information is lost. We plan to also decode visual attributes in future work when training the plot-to-table model.

More successful and failure case analyses are available in \Cref{sec:appendix_more_cases}.

\subsection{Out-of-distribution Analysis}\label{sec:ood}

One limitation of our evaluation setup is that the kind and style of charts that are part of \model{}'s training corpus are in the same domain as those in the evaluation sets from ChartQA and PlotQA.
This raises the question of whether \model{} will generalize to charts sourced from different websites or built using completely different tools.
However, few public resources exist containing both charts and their associated tables. 

In order to estimate the out-of-distribution capabilities of \model{} we annotate 10 charts from the recently released TaTa dataset~\cite{gehrmann2022tata}, sourced from \url{dhsprogram.com}. We skip choropleth maps since none have been seen during training. We find \model{} obtains an average 78\% $\texttt{RMS}_{\text{F1}}$ score in reconstructing the underlying tables. 
We observed two limitations in \model{} which we outline below and can attributed to the nature of the training datasets used. 
First the model could get distracted by adjacent text, such as references to external sources, and it benefited from cropping the chart in advance.
Secondly, \model{} struggled to understand labels or values linked to their corresponding bar/pie section by an arrow.
We will address these issues in future work by making the synthetic data creation pipeline more robust.
\section{Conclusion}\label{sec:conclusion}
We have proposed \model+LLM, a method for visual language reasoning by decomposing the task into two steps. The first step is converting a plot into linearized table using an image-to-text Transformer model finetuned for the conversion. The second step is combining the plot-to-text model with an off-the-shelf LLM to reason on the linearized table with just one-shot supervision. 

We standardize the plot-to-table conversion task by proposing a new table similarity comparison metric that considers the structure of the table and the numeric values but is invariant to column/row permutation. With the new metric, we compare our image-to-text model \model's performance with an OCR-based baseline and three end-to-end baselines, achieving the best improvement. The conversion model is then used for downstream tasks of ChartQA and PlotQA. On ChartQA human-query set, the one-shot \model+LLM model achieves +29.4\% performance compared with end-to-end SOTA finetuned with thousands of examples. We have also conducted comprehensive analyses to understand the wins and loses of the \model+LLM framework and highlight that encoding visual attributes can be a fruitful direction for future exploration.

\section*{Limitations}\label{sec:limitations}
\model{}'s strength is highly dependent on the accuracy of plot-to-text(table) conversion. To obtain effective plot-to-text conversion, large amounts of diverse and in-domain plot-table parallel data are usually needed. It is unknown to which extent \model{} can work for out-of-domain (OOD) plot-to-text conversion. We investigated this in section \Cref{sec:ood} but in the future a wider range of web charts can be used to gain a deeper understanding into \model{}'s robustness for OOD plots.

Beyond, \model{} does not work for visual language that does not have a clear latent textual representation such as textbook figures where the visual illustrations are created using specialized software and do not have clear structured representations.

Another limitation of the current \model{} approach is that we ignore any layout information such as orientation and color of the visual elements/objects. In future work, we can incorporate such attributional information by including them in the decoding target.

\section*{Ethics Statement}\label{sec:ethics}
To the best of our knowledge, \model{} is of low risk to the society since it is an information extraction model that converts graphics information from image to textual information in the form of table. That said, when combined with LLMs, \model+LLM can demonstrate potential risk such as generating toxic content similar to when LLMs are used standalone. As a result, we should proceed with caution when deploying \model+LLM in the real-world and take necessary precautions such as having a filtering stage after the generation.

In terms of data used, all training and evaluation data are either synthetically created using rules or publicly available data on the web with appropriate permissive licenses.

\bibliographystyle{acl_natbib}
\bibliography{custom,anthology}

\clearpage

\appendix
\label{sec:appendix}


\section{Details of Baselines}\label{sec:baselines}
We introduce below the details of the baselines used in \Cref{tab:main}.

T5 is an encode-decoder Transformer model proposed by \citet{raffel2020exploring}. The baseline model T5 takes the concatenation of a linearized table (and a query, when the task is QA) as input, and aims to decode the target (answer or summarization). When the gold table is availible, the gold table is used as the input and the chart image is not used directly. VL-T5 proposed by \citet{cho2021unifying} is similar to T5 but also takes a visual input (i.e., the chart image) on the encoder side. VisionTaPas \citep{masry-etal-2022-chartqa} is modified from TaPas \citep{herzig-etal-2020-tapas} to incorporate the visual modality by adding a ViT model \citep{dosovitskiy2021image} and cross-modal fusion layers.
T5-OCR, VL-T5-OCR, and VisionTaPas-OCR are the same model as T5, VL-T5, and VisionTaPas, respectively. However, they do not assume the existence of gold table but use an OCR-based system to extract the data table from the chart image.
The above mentioned models and their performance numbers are all extracted from \citet{masry-etal-2022-chartqa} and \citet{kantharaj-etal-2022-chart}. Please see the original paper for more details.
Classification - Regression Chart Transformer (CRCT) \citep{levy2022classification} is the best performing model on PlotQA according to the PlotQA benchmark on \url{paperswithcode.com}. It uses a detector that extracts all textual and visual elements of chart then processes these elements with a multimodal Transformer.
PaLI \citep{chen2022pali} with 17B parameters is a SOTA on multiple vision-language tasks in the natural image domain however fails significantly on chart understanding tasks.
MatCha \citet{liu2022matcha} is the strongest supervised baseline and uses a mixture of image-to-text tasks as pretraining to inject math reasoning and chart layout understanding knowledge to the base model.
In downstream tasks ChartQA and PlotQA, the full-supervised models are finetuned with the corresponding training sets (ChartQA has $\sim$33k data points and PlotQA has $\sim$37M).
The fully supervised results are collected from \citet{liu2022matcha}.


\section{Human Evaluation Questions}\label{sec:human_eval_form}
We list below (\Cref{fig:metric_eval}) the annotation form of the three questions asked when producing the human judgment scores of plot-table pairs. Each question asks one aspect regarding the quality of the generated table and the annotator needs to rate the table from 1 to 5.
The final table score is the average score from the three questions.

\begin{figure*}\small
\begin{lstlisting}
[plot]

[generated table]

Questions (rate 1 to 5; the higher the better):

1. Does the model overgenerate columns/rows or some rows/columns are missing?  (1 means lots of extra or missing columns/rows, 5 means no extra columns/rows and all necessary columns/rows are presented)
Answer:

2. Are the x, y label/index names, and title correct? (1 means nothing is accurate, 5 means all information are exactly correct)
Answer: 

3. Are numbers close to the true values and associated with the correct column, row labels/indexes? (1 means nothing is accurate, 5 means all values and their associated labels/indexes are accurate)
Answer:

\end{lstlisting}
\caption{Questions for producing human scores of plot-table pairs.}
\label{fig:metric_eval}
\end{figure*}

\section{Chain-of-thoughts and Program-of-thoughts Prompt}
\label{sec:cot_prompt}


In \Cref{fig:qa_prompt} we show the one-shot prompt used across all experiments CoT. It is taken from development set examples in combination with prompts used by~\citet{chen-2023-large}. We also modified this prompt to output Python code when needing to compute arithmetic operations in \Cref{fig:qa_pot_prompt}.




\begin{figure*}\small
\begin{lstlisting}
Read the table below to answer the following questions.

Header: Year | Democrats | Republicans | Independents
Row 1: 2004 | 68.1% | 45.0% | 53.0%
Row 2: 2006 | 58.0% | 42.0% | 53.0%
Row 3: 2007 | 59.0% | 38.0% | 45.0% 
Row 4: 2009 | 72.0% | 49.0% | 60.0%
Row 5: 2011 | 71.0% | 51.2% | 58.0% 
Row 6: 2012 | 70.0% | 48.0% | 53.0% 
Row 7: 2013 | 72.0% | 41.0% | 60.0%

Q: In which year republicans have the lowest favor rate?
A: Let's find the column of republicans. Then let's extract the favor rates, they
[45.0, 42.0, 38.0, 49.0, 51.2, 48.0, 41.0]. The smallest number is 38.0, that's
Row 3.  Row 3 is year 2007. The answer is 2007.

Q: What is the sum of Democrats' favor rates of 2004, 2012, and 2013?
A: Let's find the rows of years 2004, 2012, and 2013. We find Row 1, 6, 7. The
favor dates of Democrats on that 3 rows are 68.1, 70.0, and 72.0.
68.1+70.0+72=210.1. The answer is 210.1.

Q: By how many points do Independents surpass Republicans in the year of 2011?
A: Let's find the row with year = 2011. We find Row 5. We extract Independents
and Republicans' numbers. They are 58.0 and 51.2. 58.0-51.2=6.8. The answer is
6.8.

Q: Which group has the overall worst performance?
A: Let's sample a couple of years. In Row 1, year 2004, we find Republicans
having the lowest favor rate 45.0 (since 45.0<68.1, 45.0<53.0). In year 2006, Row
2, we find Republicans having the lowest favor rate 42.0 (42.0<58.0, 42.0<53.0).
The trend continues to other years. The answer is Republicans.

Q: Which party has the second highest favor rates in 2007?
A: Let's find the row of year 2007, that's Row 3. Let's extract the numbers on
Row 3: [59.0, 38.0, 45.0]. 45.0 is the second highest. 45.0 is the number of
Independents. The answer is Independents.
\end{lstlisting}
\caption{Prompt used for question answering on tables.}
\label{fig:qa_prompt}
\end{figure*}

\begin{figure*}\small
\begin{lstlisting}
Read the table below and write code to answer the following questions using the variable ans.

Header: Year | Democrats | Republicans | Independents
Row 1: 2004 | 68.1% | 45.0% | 53.0%
Row 2: 2006 | 58.0% | 42.0% | 53.0%
Row 3: 2007 | 59.0% | 38.0% | 45.0% 
Row 4: 2009 | 72.0% | 49.0% | 60.0%
Row 5: 2011 | 71.0% | 51.2% | 58.0% 
Row 6: 2012 | 70.0% | 48.0% | 53.0% 
Row 7: 2013 | 72.0% | 41.0% | 60.0%

Q: What was the average difference in approval rates between democrats and republicans in 2006 and 2007?
#Python
democrats_2006 = 58.0
republicans_2006 = 42.0
difference_2006 = democrats_2006 - republicans_2006
democrats_2007 = 59.0
republicans_2007 = 38.0
difference_2007 = democrats_2007 - republicans_2007
ans = (difference_2006 + difference_2007) / 2

Q: What is the average of Democrats' favor rates of 2004, 2012, and 2013?
#Python
# Years 2004, 2012, and 2013  correspond to rows 1, 6 and 7.
democrats_2004 = 68.1
democrats_2012 = 70.0
democrats_2013 = 72.0
ans = (democrats_2004 + democrats_2012 + democrats_2013) / 3

Q: Which party had less than 50% approval rate in 2013?
#Python
# year 2013 corresponds to row 7. Numbers on row 7 are [72.0, 41.0, 60.0]
# Republicans are the only with less than 50.
ans = "Republicans"

Q: What is percentage of relative increase in approval rate for democrats from 2012 to 2013?
#Python
# year 2012 and 2013 correspond to rows 6 and 7.
# Numbers on row 6 are [70.0, 48.0, 53.0]
# Numbers on row 7 are [72.0, 41.0, 60.0]
democrats_2012 = 70.0
democrats_2013 = 72.0
ans = 100 * (democrats_2013 - democrats_2012) / democrats_2012

Q: What is the difference between republicans in 2011 and democrats in 2006?
#Python
# year = 2011 corresponds to row 5 and the republicans had a 51.2 rate
republicans_2011 = 51.2
# year = 2006 corresponds to row 2 and the democrats had a 58.0 rate
democrats_2006 = 58.0
# The difference between A and B is A - B which may be negative
ans = republicans_2011 - democrats_2006
\end{lstlisting}
\caption{Prompt used for question answering on tables using Python code.}
\label{fig:qa_pot_prompt}
\end{figure*}

\section{More Case Study}\label{sec:appendix_more_cases}

\paragraph{Successes.}
In \Cref{tab:case_study_x1} and \Cref{tab:case_study_x2} we demonstrate two more cases where \model+LLM are successful due to its stronger numerical reasoning capabilities.

\begin{table}[ht!]
  \centering
  \small
  \begin{tabular}{c}
    \begin{minipage}{.46\textwidth}
      \frame{\includegraphics[width=\linewidth, height=56mm]{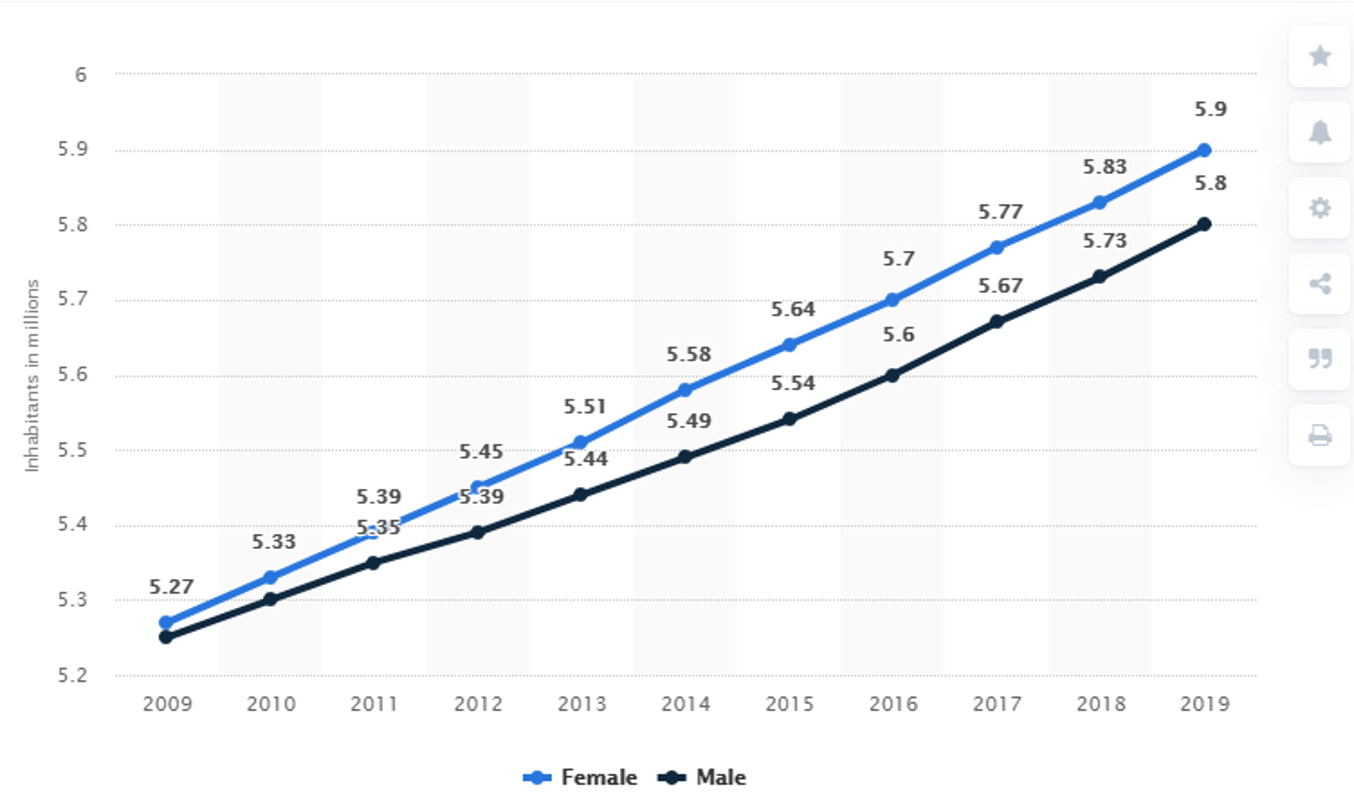}}
    \end{minipage} 
    \\
    \\
    \begin{minipage}[t]{.46\textwidth}
    \textit{\textit{Question:} Can you identify how many points the female data moved from 2009 to 2019?} \\ \\
    \textsc{MatCha}: \textcolor{red}{\textbf{0.8}} \\ \\
    \model+FlanPaLM CoT: Let's find the row of year 2009, that's Row 11. Let's extract the number on Row 11, that's 5.27. Let's find the row of year 2019, that's Row 1. Let's extract the number on Row 1, that's 5.9. 5.9-5.27=0.63. The answer is \textcolor[rgb]{0.4,0.8,0}{\textbf{0.63}}.\\ \\
    \model+Codex PoT:    \begin{lstlisting}[language=Python]
#Python
#year 2009 corresponds to row 11
#year 2019 corresponds to row 1
female_2009 = 5.27 
female_2019 = 5.9 
ans = female_2019 - female_2009 
    \end{lstlisting}
Compiler output: \textcolor[rgb]{0.4,0.8,0}{\textbf{0.63}}
    \end{minipage}
  \end{tabular}
  \caption{A line plot example that requires numerical reasoning which is easily solved by \model+LLM but failed by \textsc{MatCha}.}\label{tab:case_study_x1}
\end{table}

\begin{table}[ht!]
  \centering
  \small
  \begin{tabular}{c}
    \begin{minipage}{.46\textwidth}
      \frame{\includegraphics[width=\linewidth, height=56mm]{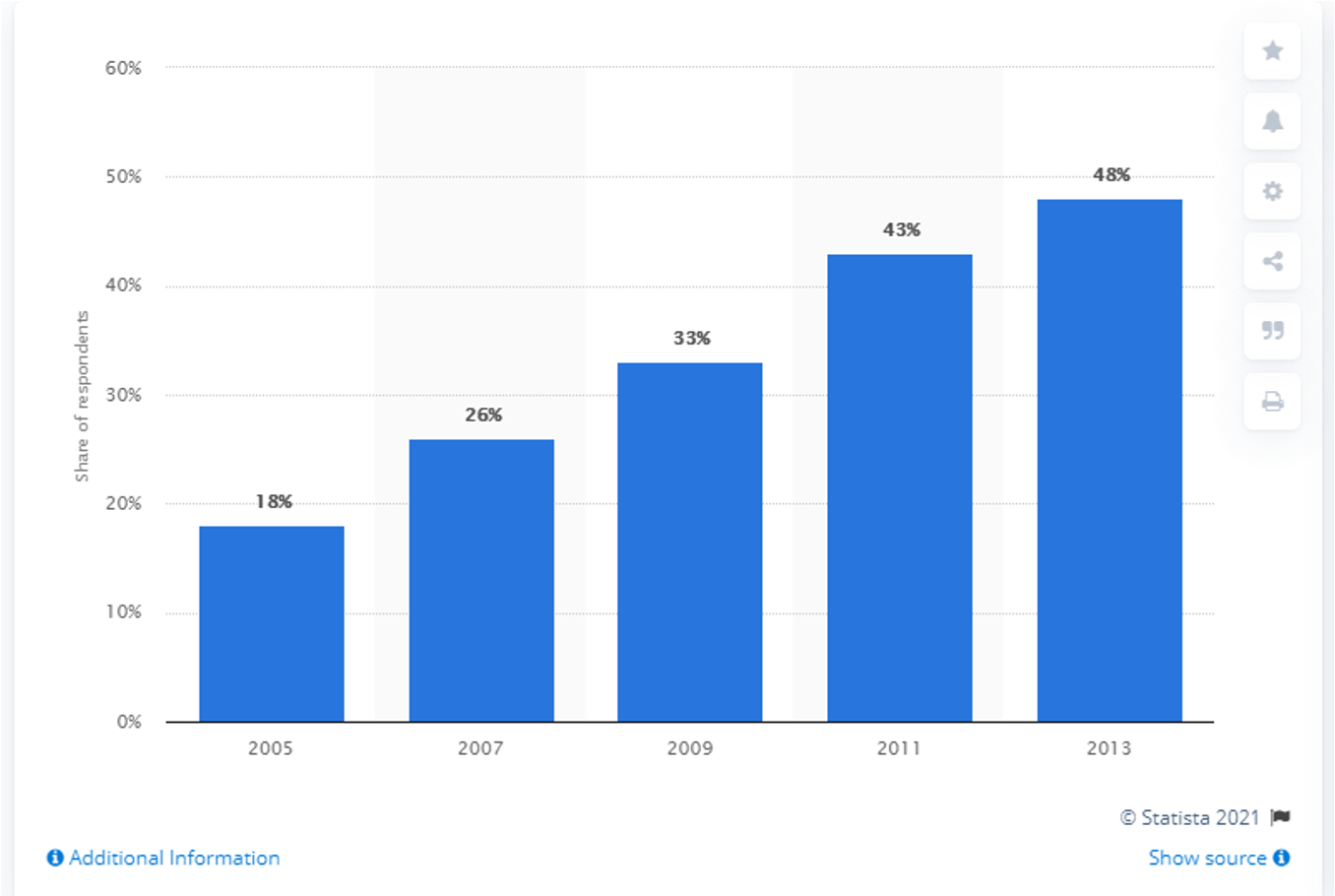}}
    \end{minipage} 
    \\
    \\
    \begin{minipage}[t]{.46\textwidth}
    \textit{\textit{Question:} What is the average penetration rate?} \\ \\
    \textsc{MatCha}: \textcolor{red}{\textbf{43}} \\ \\
    \model+FlanPaLM CoT: Let's add all the penetration rates: 48\% + 43\% + 33\% + 26\% + 18\% = 168\%. There are 5 years, so the average is 168\% / 5 = 33.6\%. The answer is \textcolor[rgb]{0.4,0.8,0}{\textbf{33.6}}.\\ \\
    \model+Codex PoT:    \begin{lstlisting}[language=Python]
#Python
# Years 2013, 2011, 2009, 2007, and 2005 correspond to rows 1, 2, 3, 4, and 5.
penetration_2013 = 48
penetration_2011 = 43
penetration_2009 = 33
penetration_2007 = 26
penetration_2005 = 18
ans = (penetration_2013 + penetration_2011 + penetration_2009 + penetration_2007 + penetration_2005) / 5
    \end{lstlisting}
Compiler output: \textcolor[rgb]{0.4,0.8,0}{\textbf{33.6}}
    \end{minipage}
  \end{tabular}
  \caption{\model+LLM is especially suitable for more complex numerical computations such as computing the average for multiple numbers as shown in this example.}\label{tab:case_study_x2}
\end{table}

\paragraph{Failures.}
For questions concerning color or other visual attributes of the graph, the \model+LLM framework is unable to handle since such information is lost in modality translation and not considered in the current textual table encoding scheme. We show an additional examples in \Cref{tab:error_x1}. 

Besides color, plot-to-table conversion can ignore other visual attributes such as the example in \Cref{tab:error_x2}. There does not exist a one-to-one alignment between dots on the line graphs and x labels. The \model{} model produces a table with only x labels and the extrapolated y values and ignore the dots in the graph.

\begin{table}[ht!]
  \centering
  \small
  \begin{tabular}{c}
    \begin{minipage}{.46\textwidth}
      \frame{\includegraphics[width=\linewidth, height=76mm]{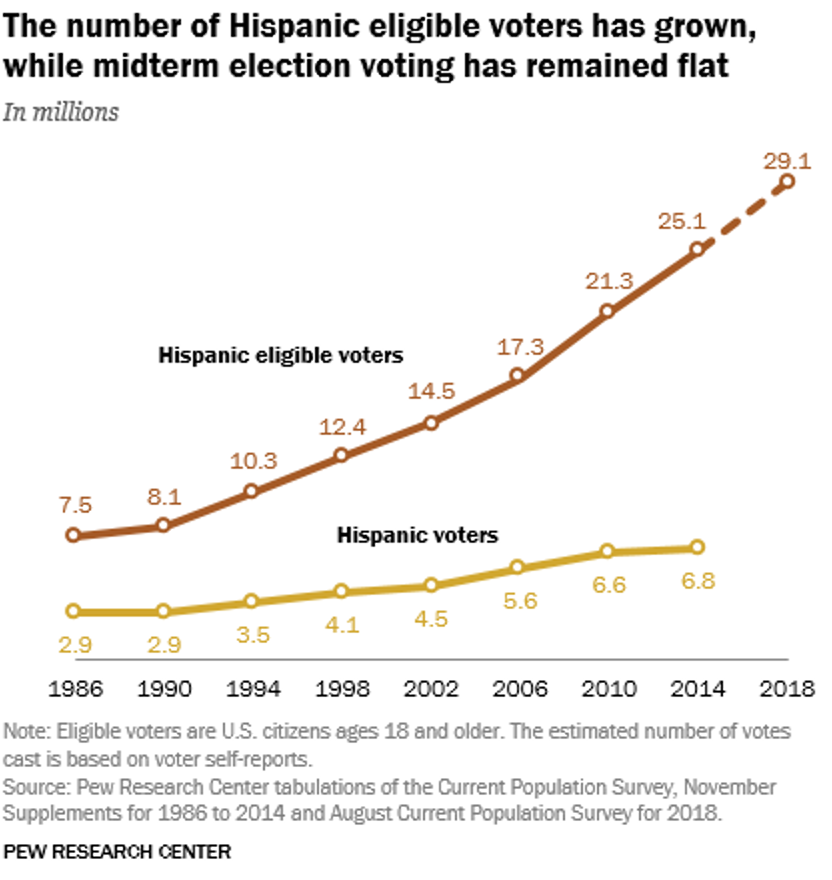}}
    \end{minipage} 
    \\
    \\
    \begin{minipage}[t]{.46\textwidth}
    \textit{\textit{Question:} What's the median value of the red graph?} \\ \\
  \textsc{MatCha}: \textcolor{red}{\textbf{10.3}} \\ \\
  \model+FlanPaLM CoT: Let's find the median value of red graph. The median is the middle value of the data set. The data set is [2, 2, 3.5, 0, 4.5, 5.6, 6.6, 6.8, 0]. The middle value is 4.5. The answer is \textcolor{red}{\textbf{4.5}}. \\ \\
\model+Codex PoT:
\begin{lstlisting}[language=Python]
#Python
# The median of the red graph is the number in the middle of the sorted list of values.
# The values are 2, 2, 3.5, 0, 4.5, 5.6, 6.6, 6.8, 0.
# The median is 4.5.
ans = 4.5
\end{lstlisting}
Compiler output: \textcolor{red}{\textbf{4.5}} 
\end{minipage}
  \end{tabular}
  \caption{All models made mistakes. For the finetuned model \textsc{MatCha}, the error seems to be numerical reasoning since its 10.3 is from the correct graph. However, for \model+LLMs, the error originates from identifying the wrong graph as the red graph.}\label{tab:error_x1}
\end{table}

\begin{table*}[ht!]
  \centering
  \small
  \begin{tabular}{c}
    \begin{minipage}{.70\textwidth}
      \frame{\includegraphics[width=\linewidth, height=76mm]{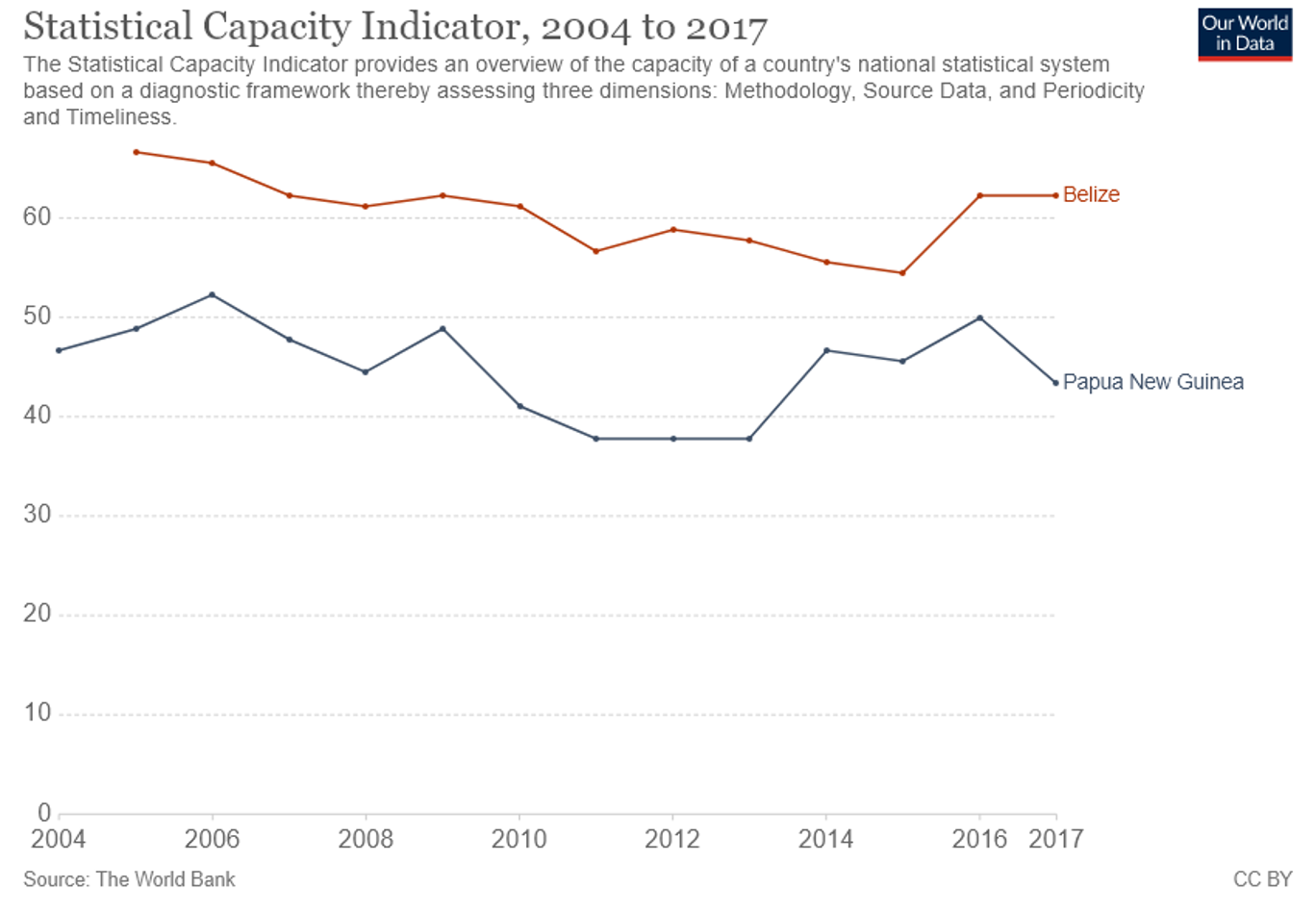}}
    \end{minipage} 
    \\
    \\
    \begin{minipage}[t]{.96\textwidth}
    \textit{\textit{Question:} Which country's line bar shows three data points in the same place between 2010 and 2014?} \\ \\
  \textsc{MatCha}: \textcolor{red}{\textbf{Belize}} \\ \\
  \model: 
  
  \texttt{\small
  Entity | 2004 | 2006 | 2008 | 2010 | 2012 | 2014 | 2016 | 2017 \\
  Belize | 65.0 | 65.0 | 61.0 | 61.0 | 58.0 | 55.0 | 62.0 | 62.0 \\
  Papua New Guinea | 44.0 | 52.0 | 44.0 | 41.0 | 37.0 | 46.0 | 47.0 | 43.0
  }\\
  \\
  \model+FlanPaLM CoT: Let's find the rows with countries. We find Row 1, 2. Let's find the years between 2010 to 2014. We find Row 6, 7. We find the data in the same place between 2010 to 2014. We find 61.0, 61.0, 41.0, 41.0. The answer is \textcolor{red}{\textbf{Belize}}. \\ \\
\model+Codex PoT:
\begin{lstlisting}[language=Python]
#Python
# Row 1 corresponds to Belize. Numbers on row 1 are [65.0, 65.0, 61.0, 61.0, 58.0, 55.0, 62.0, 62.0]
# Row 2 corresponds to Papua New Guinea. Numbers on row 2 are [44.0, 52.0, 44.0, 41.0, 37.0, 46.0, 47.0, 43.0]
# Belize has three data in the same place between 2010 to 2014.
ans = "Belize"
\end{lstlisting}
Compiler output: \textcolor{red}{\textbf{Belize}} 
\end{minipage}

  \end{tabular}
  \caption{An error caused by plot-to-table translation. The dots on the lines and the x labels (years) are not exactly aligned, causing challenges in the translation.}\label{tab:error_x2}
\end{table*}

\end{document}